\documentclass{article} 
\usepackage{inconsolata}
\usepackage[preprint]{colm2025_conference}

\usepackage{microtype}
\usepackage{hyperref}
\usepackage{url}
\usepackage{xurl}
\usepackage{breakurl}
\usepackage{booktabs}
\usepackage{graphicx}
\usepackage{subcaption}
\usepackage{lineno}
\usepackage{multirow}
\usepackage{booktabs} 
\usepackage[utf8]{inputenc}
\usepackage{tcolorbox}

\definecolor{darkblue}{rgb}{0, 0, 0.5}
\hypersetup{colorlinks=true, citecolor=darkblue, linkcolor=darkblue, urlcolor=darkblue}

\title{It's Not That Simple. An Analysis of Simple Test-Time Scaling}

\author{
Guojun Wu \\
Independent Researcher \\
Ningbo, China \\
\texttt{guojun\_wu@outlook.com}
}

\begin{document}

\ifcolmsubmission
\linenumbers
\fi

\maketitle

\begin{abstract}
Prior work proposed simple test-time scaling, a method for replicating this scaling behavior with models distilled from o1-like models by manually controlling test-time compute: either scaling down by enforcing a maximum length or scaling up by iteratively appending “Wait” when the model is about to terminate its generation. This paper presents an analysis of simple test-time scaling and finds that the scaling behavior is largely attributed to scaling down by enforcing a maximum length. In contrast, fine-tuning on long CoT data distilled from o1-like models has no significant impact on scaling behavior, and scaling up by appending “Wait” leads to inconsistencies, as the model may oscillate between solutions. A key distinction exists between scaling down by enforcing a maximum length and scaling up test-time compute in o1-like models, such as DeepSeek-R1. These models are typically allowed to utilize as much compute as needed, with the only constraint being the model’s maximum supported length. By learning to naturally scale up test-time compute during reinforcement learning, o1-like models surpass their peak performance when scaling up. In contrast, simple test-time scaling progressively imposes a lower upper limit on model performance as it scales down. While replicating the test-time scaling behavior of o1 models can be straightforward by scaling down, it is crucial to recognize that the goal of scaling test-time compute is to unlock higher performance—beyond what the model could originally achieve—rather than merely reproducing the appearance of scaling behavior.
\end{abstract}
\begin{figure}[hb]
    \centering
    \begin{subfigure}[b]{0.48\textwidth}
        \includegraphics[width=\textwidth]{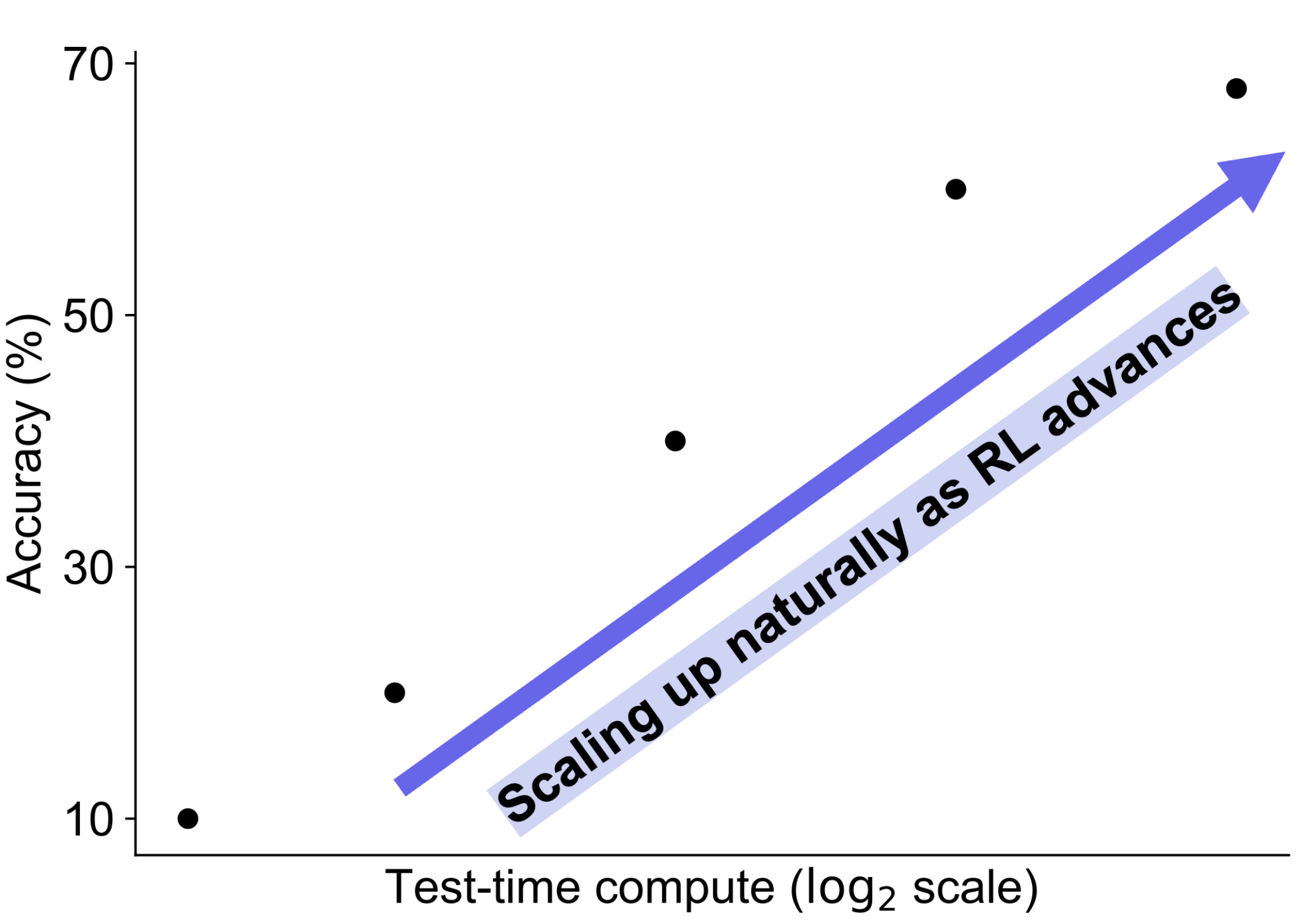} 
        \caption{DeepSeek-R1-Zero}
        \label{fig:up}
    \end{subfigure}
    \hfill
    \begin{subfigure}[b]{0.48\textwidth}
        \includegraphics[width=\textwidth]{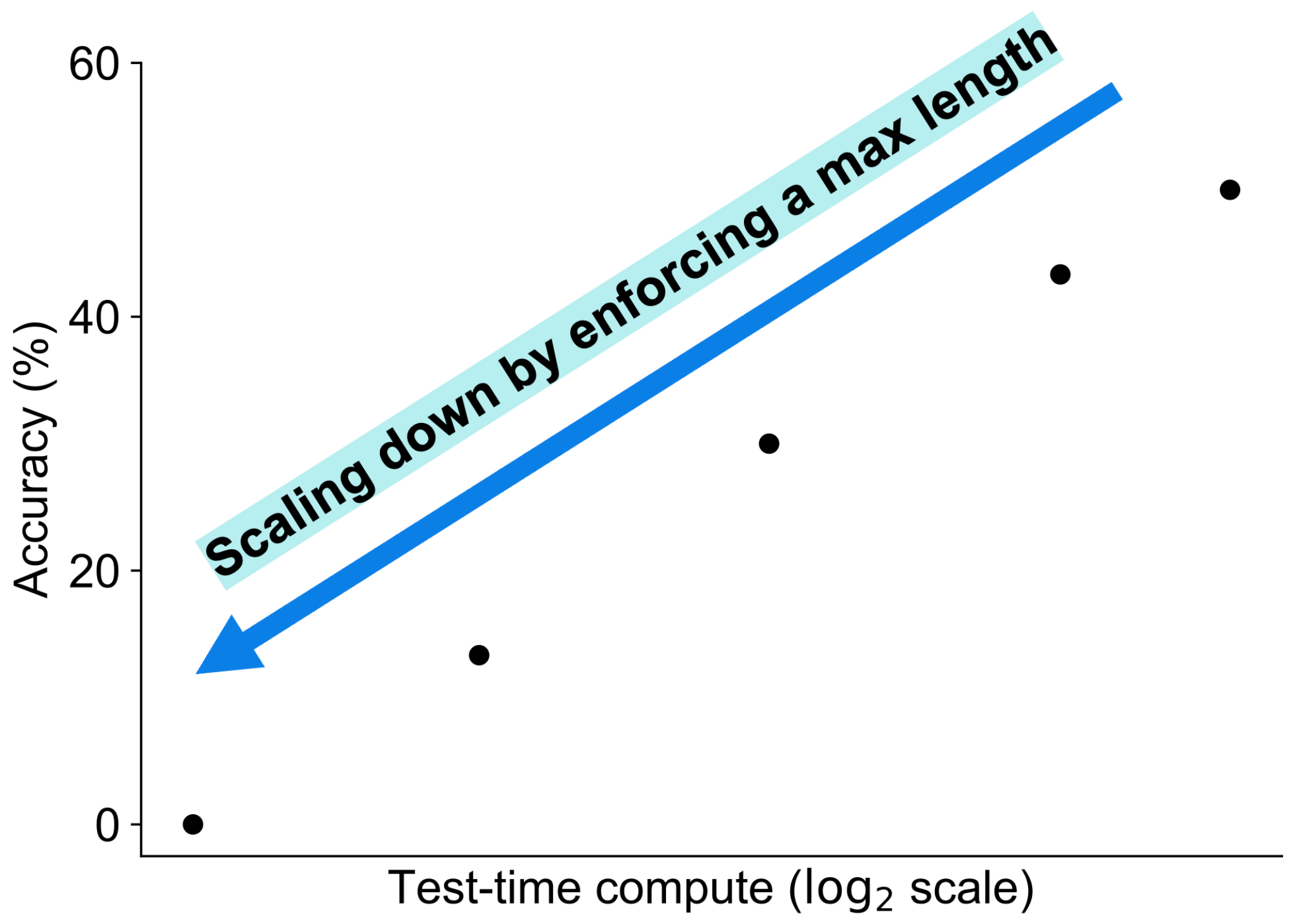} 
        \caption{s1}
        \label{fig:down}
    \end{subfigure}
    \caption{Two distinct causes lead to similar test-time scaling behavior. DeepSeek-R1-Zero naturally learns to scale up test-time compute during reinforcement learning, leading to improved performance, as illustrated in (a). In contrast, s1 exhibits a similar behavior, as shown in (b), but this is primarily due to enforcing a maximum length on the generation process. As the maximum length is set to a lower value, performance degrades.}
    \label{fig:up_down}
\end{figure}

\section{Introduction}
Recently, there has been a paradigm shift in scaling compute for large language models (LLMs), transitioning from scaling pre-training \citep{NEURIPS2020_1457c0d6,touvron2023llamaopenefficientfoundation,bai2023qwentechnicalreport} to scaling test-time compute \citep{feng2024alphazeroliketreesearchguidelarge,snell2024scalingllmtesttimecompute,xin2024deepseekproverv15harnessingproofassistant}. OpenAI’s o1 \citep{openai2024reasoning} models exemplify this shift, demonstrating strong performance across various reasoning tasks, such as mathematics \citep{hendrycks2021measuringmathematicalproblemsolving} and software engineering \citep{jimenez2024swebench}, with consistent improvements as test-time compute increases. A distinguishing feature of these models is their ability to scale inference compute through extended chains of thought (CoTs) \citep{nye2021workscratchpadsintermediatecomputation,wei2023chainofthoughtpromptingelicitsreasoning}. 

Several efforts \citep{kimiteam2025kimik15scalingreinforcement,deepseekai2025deepseekr1incentivizingreasoningcapability} have attempted to replicate the performance of o1 models by training LLMs to scale up test-time compute by increasing the length of the CoT reasoning process. DeepSeek-R1 demonstrates that LLMs can naturally increase test-time compute and show consistent performance improvements through a pure reinforcement learning (RL) process, ultimately achieving performance on par with o1. 

Among the replication efforts, \cite{muennighoff2025s1simpletesttimescaling} aim to identify the simplest approach, referred to as simple test-time scaling, to achieve test-time scaling. They have shown that fine-tuning a model with just 1,000 questions paired with reasoning traces distilled from an o1-like model \citep{google2024gemini} leads to the s1 model, which can achieve strong reasoning performance. Additionally, they introduce a test-time intervention technique called budget forcing, which controls test-time compute by terminating the model’s thinking process at a certain length or extending it by appending “Wait” multiple times to the model’s generation when it attempts to end. With budget forcing, their s1 model demonstrates clear test-time scaling behavior.

However, the mechanisms underlying the test-time scaling behavior remain unclear, as multiple components are involved. In short, simple test-time scaling relies on three key components: (1) fine-tuning on long CoT data distilled from o1-like models, (2) scaling down by enforcing a maximum length, and (3) scaling up by repeatedly appending “Wait” to the model’s generation when it attempts to terminate.

\begin{figure}[t]
\begin{center}
\includegraphics[width=1\linewidth]{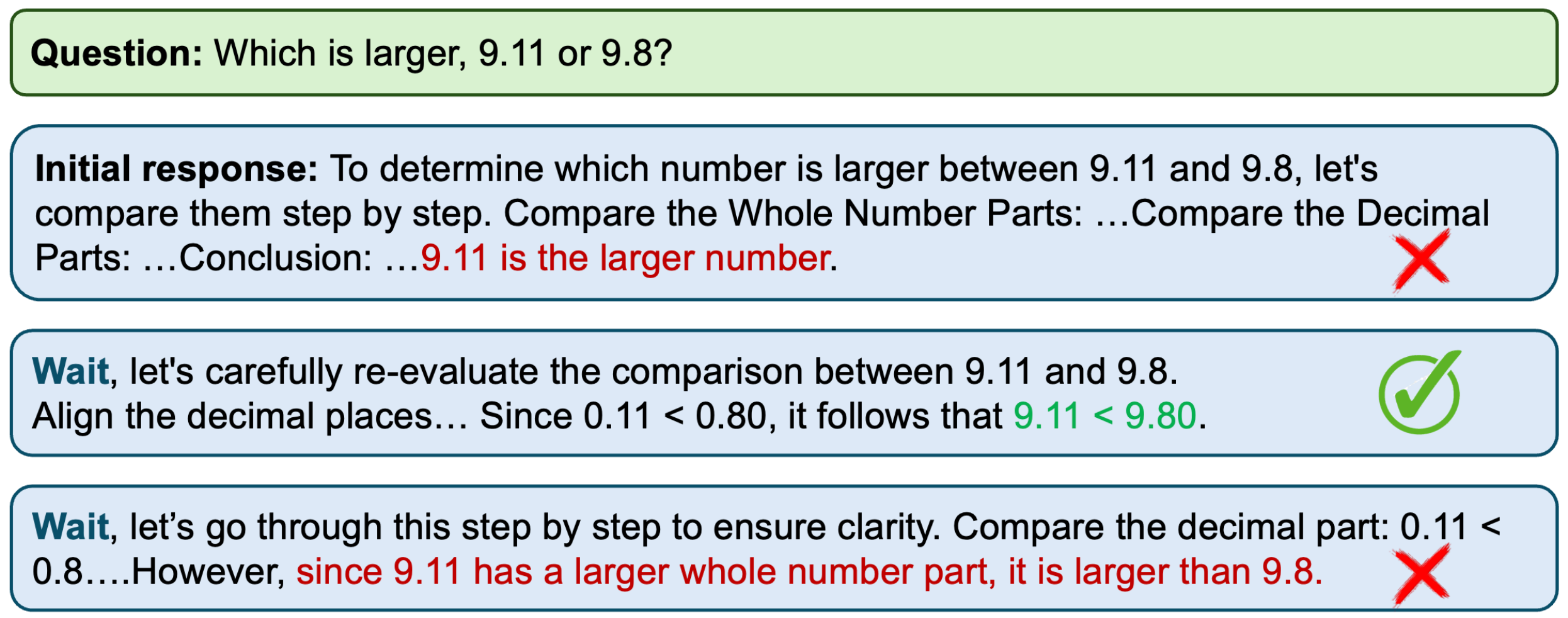}
\end{center}
\caption{An example of the model changing its answer to a question as "Wait" is appended iteratively illustrates that the answer can oscillate between correct and incorrect.}
\label{fig:flip}
\end{figure}

In this work, we examine these ingredients separately. We involve more models that have been fine-tuned on long CoT data distilled from o1-like models and others that have not. Our findings show that the models can exhibit test-time scaling behavior regardless of whether they have been distilled. Furthermore, we identify that scaling down by forcing a maximum length is the primary — if not the only — contributor to the test-time scaling behavior. As the maximum length is gradually reduced, performance degrades. When scaling up by appending “Wait” multiple times, we observe inconsistent performance. For instance, the performance of s1 increases when appending “Wait” for the fourth time but decreases on the fifth. This inconsistency is caused by the model oscillating between answers on the same problem, as illustrated in Figure \ref{fig:flip}. Moreover, appending “Wait” is inefficient, as the answers or even the entire solution often remain unchanged after appending. We also demonstrate that there may be no easy fix for this inefficiency.

Based on our empirical results, we emphasize that while similar test-time scaling behavior may appear on the surface, there are key differences in how such behavior is achieved. As shown in Figure \ref{fig:up_down}, for the o1-like model DeepSeek-R1-Zero, this behavior naturally emerges as RL advances, with the model continually striving to achieve its best performance on the problems. In contrast, for s1, the maximum length is manually controlled, meaning that problems requiring longer reasoning will be prematurely answered, setting an upper limit on the model's performance. As the maximum length is reduced, more problems are answered prematurely, and performance gradually decreases to zero when the maximum length becomes too short.

We highlight that it is simple to obtain test-time scaling behavior by manually scaling down, but it is crucial to understand that the purpose of scaling test-time compute is to achieve better performance than what the models can initially provide. The test-time scaling behavior, to some extent, is a by-product of this pursuit. Simply pursuing superficial patterns can be distracting at best and misleading at worst.

\section{Related Work}
We discuss both manual and natural approaches to scaling test-time compute. Manual methods require explicitly configuring hyperparameters to control the generation process, whereas natural approaches allow models to autonomously scale their compute.

\subsection{Scaling test-time compute manually }
\noindent\textbf{Scaling in parallel.} Previous works have scaled test-time computation by generating multiple solution attempts in parallel and selecting the best outcome based on predefined criteria. These criteria include choosing the most frequent response for majority voting \citep{wang2023selfconsistencyimproveschainthought} or selecting the highest-scoring response based on an external reward in Best-of-N methods \citep{irvine2023rewardingchatbotsrealworldengagement,brown2024largelanguagemonkeysscaling}. More recently, studies \citep{wang2025thoughtsplaceunderthinkingo1like} have observed that incorrect model outputs tend to be longer, while correct ones are often shorter, suggesting that a Shortest-of-N \footnote{\url{https://x.com/AlexGDimakis/status/1885447830120362099}} approach can be an effective alternative.

\noindent\textbf{Scaling sequentially.} Unlike parallel sampling, sequential scaling methods generate answers iteratively, incorporating feedback from previous responses to refine the output. \cite{stechly2023gpt4doesntknowits} explores iterative prompting strategies to improve the accuracy of LLMs on reasoning tasks, using either the models themselves or an external, provably correct verifier to provide feedback. Furthermore, \cite{stechly2024self} proposes the LLM-Module framework, where multiple verifiers offer feedback to guide the refinement process. Additionally, a simple form of sequential scaling \citep{muennighoff2025s1simpletesttimescaling} involves appending a cue (e.g., "Wait"), prompting the model to continue reasoning and engage in reflection, albeit without explicit feedback.

\noindent\textbf{Scaling with a hybrid approach.} Tree-based search methods \citep{gandhi2024streamsearchsoslearning, wu2025inferencescalinglawsempirical} provide a hybrid approach between sequential and parallel scaling, incorporating techniques such as Monte Carlo Tree Search (MCTS) \citep{zhang2023planninglargelanguagemodels,liu2024dontthrowawayvalue,choi2023kctsknowledgeconstrainedtreesearch} and guided beam search \citep{xie2023selfevaluationguidedbeamsearch}. \cite{snell2024scalingllmtesttimecompute} proposes a combination of parallel sampling and sequential revision, where multiple sequential chains are generated in parallel, and the best answer is selected across the chains.

\subsection{Scaling test-time compute naturally}
\noindent\textbf{Reinforcement learning.} With DeepSeek-R1 \citep{deepseekai2025deepseekr1incentivizingreasoningcapability} being the first to achieve performance on par with o1, it has demonstrated that models can naturally learn to scale test-time computation through reinforcement learning (RL) with outcome rewards. Several follow-up studies \citep{zeng2025simplerl, yeo2025demystifyinglongchainofthoughtreasoning, deepscaler2025} have also found that the simple RL process can be effective for smaller models, enabling them to scale up test-time compute with improved performance.

\noindent\textbf{Distillation.} \cite{deepseekai2025deepseekr1incentivizingreasoningcapability} has shown that small models can achieve impressive performance and generate long CoTs by distilling DeepSeek-R1. Further work \citep{ye2025limoreasoning, muennighoff2025s1simpletesttimescaling} also finds that fine-tuning on a small sample distilled from o1-like models enables small models to produce long CoTs with competitive performance.

\section{Experiments and Results}\label{results}
Simple test-time scaling involves three key components: (1) fine-tuning on long chain-of-thought (CoT) data distilled from models similar to o1, (2) scaling down by enforcing a maximum output length, and (3) scaling up by repeatedly appending the token "Wait" when the model attempts to terminate its generation. This study investigates the contribution of each component to test-time scaling behavior. To better isolate their effects, we analyze them separately as follows:

\begin{enumerate}
\item Except for s1, we evaluate four additional models: DeepSeek-V3, Qwen2.5-32B-Instruct, Qwen2.5-72B-Instruct, and r1-distill-Qwen-32B. Among these, s1, r1-distill-Qwen-32B, and DeepSeek-V3 have been fine-tuned on long CoT data distilled from reasoning models, whereas the other two have not. Additionally, Qwen2.5-32B-Instruct serves as the base model for s1, allowing for an ablation study.
\item In Section \ref{sec:down}, we analyze the impact of scaling down by enforcing a maximum output length. For each model, we progressively reduce the maximum length. Once a model’s output surpasses this limit, it is hard-truncated and required to provide an answer without further reasoning.
\item In Section \ref{sec:up}, we examine the effect of scaling up by appending "Wait." Whenever a model is about to terminate, we append "Wait" to its output and re-prompt it with the full conversation history. We further investigate the inefficiencies observed when applying this method.
\end{enumerate}

We evaluate s1 and the four other models on the AIME 2024 benchmark. For s1, we use publicly available generation outputs. For the remaining models, we collect outputs via API. To ensure consistency, all four models are prompted using a system instruction adapted from prior work \citep{zeng2025simplerl}.

\begin{tcolorbox}[colframe=black, colback=white, coltitle=black, sharp corners=south, fonttitle=\bfseries]
\textbf{System prompt:} 
Solve the following math problem efficiently and clearly. The last line of your response should be of the following format: 'Therefore, the final answer is: \texttt{\textbackslash boxed\{\{ANSWER\}\}}' Think step by step before answering.
\end{tcolorbox}

\subsection{Scaling Down by Enforcing a Maximum Length}\label{sec:down}

\noindent\textbf{Setup.} In the original setting of s1, a maximum length is enforced by appending an end-of-thinking token delimiter, prompting the model to exit the reasoning stage early and provide its best possible answer. However, during our implementation with the other four models, we observed that simply adding a hint or an instruction at the end of the response was insufficient to make the models stop reasoning and provide an early answer.

Instead, we found that modifying the system prompt, as shown below, to explicitly instruct the model to cease reasoning and provide its best guess based on the available information was more effective. This adjustment successfully prompted most models to end their reasoning and generate an answer. However, r1-distill-Qwen-32B failed to follow this instruction and continued reasoning. As no reliable method was found to enforce a maximum length for this model, we exclude it from this investigation.

\begin{tcolorbox}[colframe=black, colback=white, coltitle=black, sharp corners=south, fonttitle=\bfseries]
\textbf{System prompt to force early answer generation:} 
Give the answer directly without any explanation or reasoning. Use this format: 'Therefore, the final answer is: \texttt{\textbackslash boxed\{\{ANSWER\}\}}' For example, 'Therefore, the final answer is: \texttt{\textbackslash boxed\{\{5\}\}}' Follow the instructions carefully.
\end{tcolorbox}

For s1, the maximum lengths are set to 500, 1k, 2k, 4k, and 8k, respectively. For DeepSeek-V3, they are 512, 1024, 2048, and 4096. For Qwen2.5-72B-Instruct, the values are 256, 512, 800, and 1024, while for Qwen2.5-32B-Instruct, they are 256, 512, and 1024.

These values are chosen based on the models' typical output lengths, ensuring that the highest value corresponds to a length that the model rarely exceeds (fewer than 10\% of cases). This setup allows most models to generate responses without interruption when using the highest length limit. We then scale down logarithmically with a factor of 2. Notably, we later introduce an additional setting of 800 for Qwen2.5-72B-Instruct, as we observe that its performance degrades too rapidly. The temperature is set to 0 for all models.

\begin{figure}[b]
    \centering
    \begin{subfigure}[b]{0.235\textwidth} 
        \includegraphics[width=\textwidth]{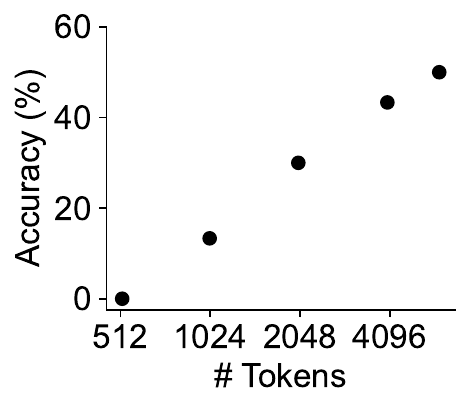}
        \caption{s1}
        \label{fig:s1}
    \end{subfigure}
    \hspace{\fill}
    \begin{subfigure}[b]{0.235\textwidth}
        \includegraphics[width=\textwidth]{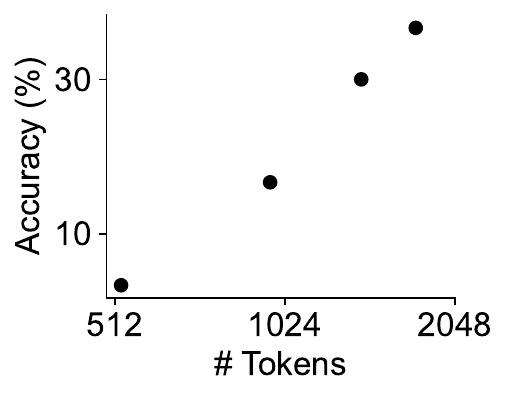}
        \caption{Deepseek-V3}
        \label{fig:deepseek}
    \end{subfigure}
    \hspace{\fill}
    \begin{subfigure}[b]{0.235\textwidth}
        \includegraphics[width=\textwidth]{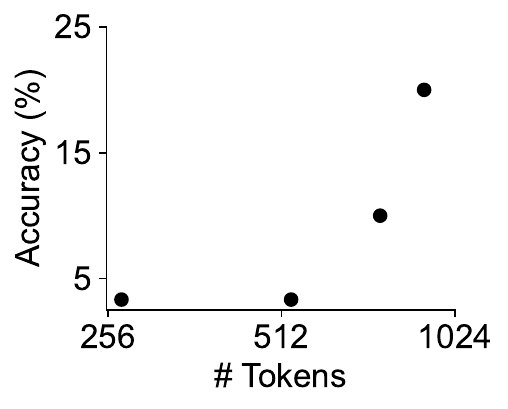}
        \caption{\small Qwen2.5-72B}
        \label{fig:72b}
    \end{subfigure}
    \hspace{\fill}
    \begin{subfigure}[b]{0.235\textwidth}
        \includegraphics[width=\textwidth]{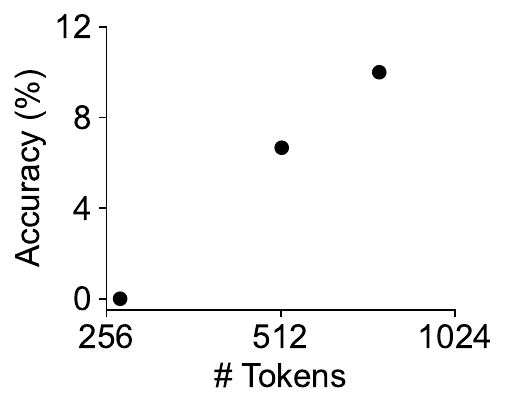}
        \caption{\small Qwen2.5-32B}
        \label{fig:32b}
    \end{subfigure}
    \caption{Scaling down by enforcing a maximum length across four models demonstrates clear test-time scaling behaviors, regardless of whether the models have been distilled with long CoT data or not.}
    \label{fig:max}
\end{figure}

\noindent\textbf{Results.} Figure \ref{fig:max} shows that when we scale down by enforcing a maximum length, test-time scaling behavior is consistently observed across all models. This indicates that fine-tuning on distilled long CoT data does not significantly impact this behavior. Furthermore, a model's reasoning ability is also irrelevant to test-time scaling. For example, Qwen2.5-32B-Instruct exhibits low accuracy even when the maximum length is set to the highest value, yet it still demonstrates a clear test-time scaling pattern.

For models that typically generate shorter outputs or are short-CoT models—such as Qwen2.5-32B-Instruct and Qwen2.5-72B-Instruct—we observe that once the maximum length is approximately 100 tokens fewer than their initial requirement, originally correct answers start becoming incorrect. This suggests that performance degradation during scaling down is largely a consequence of problems requiring different solution lengths for mature answers. Specifically, problems that can be solved with shorter responses remain unaffected for longer, while those requiring longer reasoning chains begin failing earlier.

Conversely, models that generate longer outputs or are long-CoT models exhibit greater tolerance to truncation. For instance, s1 can still produce correct answers even when only one-third of its original output remains. Previous studies \citep{chen2025think23overthinkingo1like} on long-CoT models indicate that these models often introduce redundancy in their reasoning (e.g., providing multiple solutions in a single output while reaching the correct answer early on). This redundancy explains why such models can better withstand output truncation.

Overall, enforcing a maximum length forces models to provide premature answers, leading to incorrect responses—though for long-CoT models, this effect may occur at a later stage. Notably, the maximum length effectively acts as an upper bound on model performance as we scale down.

\begin{tcolorbox}[
    colback=blue!5!white, 
    colframe=black, 
    sharp corners=south, 
    title={\textbf{Takeaways for scaling down by enforcing a maximum length}}, 
    fonttitle=\bfseries, 
    coltitle=white, 
    colbacktitle=black
]
All models, regardless of whether they have been distilled on long CoT data, exhibit clear test-time scaling behavior. The key factor influencing this behavior is the varying solution length required for different problems. Each time we scale down, a portion of previously correct problems becomes incorrect, as the models are forced to provide answers prematurely. Short-CoT models have lower tolerance for truncation, as their outputs contain less redundancy.
\end{tcolorbox}

\subsection{Scaling Up by Appending ``Wait''} \label{sec:up}
In Section \ref{sec:flip}, we begin by analyzing the model's response each time "Wait" is appended. We examine how the answer evolves and manually review the response after each "Wait." We observe inconsistent performance as "Wait" is appended, along with significant repetition in both the answers and the responses.  

To further investigate the extent of this repetition, we conduct three additional runs in Section \ref{sec:repeat}, providing a more detailed analysis of these inefficiencies. Finally, in Section \ref{sec:fix}, we explore potential quick fixes for this issue but fail to find an effective solution.

\subsubsection{Inconsistent Performance} \label{sec:flip}
\noindent\textbf{Setup.} Initially, we set the maximum length to the highest value supported by each model to ensure they can generate complete solutions. Specifically, the maximum length is set to 32,768 for s1, 8,196 for DeepSeek-V3, and 16,384 for the other three models. We observe that all models remain within these limits, except for DeepSeek-V3, which occasionally (2 out of 30 cases) exceeds the maximum length.  

For s1, we present results with seven instances of "Wait" appended, whereas for the other models, we use only two. After each "Wait" is appended, we re-prompt the model with the entire conversation history and collect its response. From these responses, we extract answers corresponding to each "Wait." The temperature is set to 0 for all models.

\begin{figure}[ht]
    \centering
    \begin{subfigure}[b]{0.9\textwidth} 
        \includegraphics[width=\textwidth]{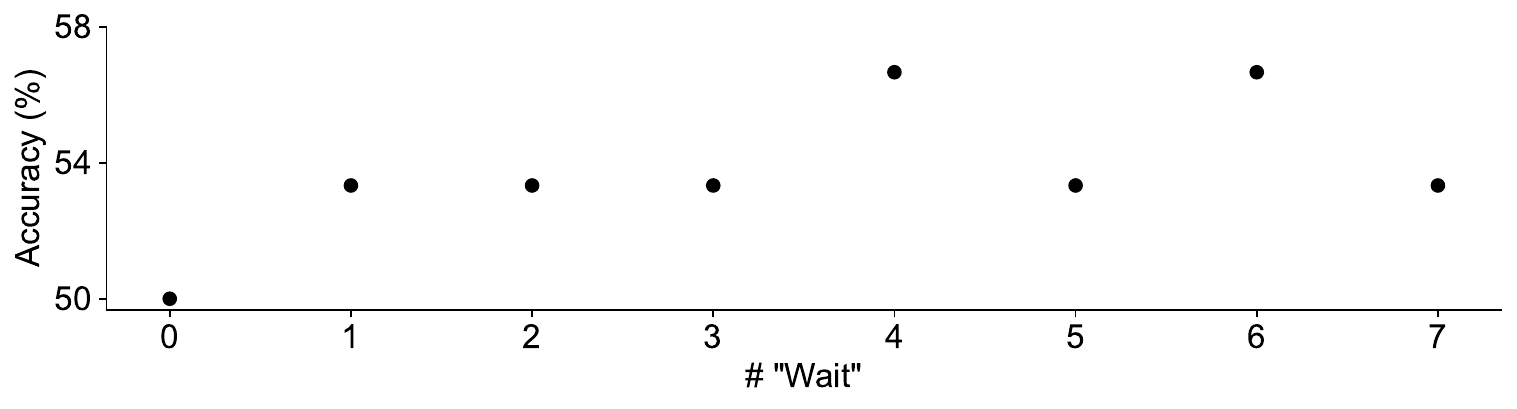}
        \caption{s1}
        
    \end{subfigure}
    
    \vspace{0.5cm} 
    
    \begin{subfigure}[b]{0.24\textwidth} 
        \includegraphics[width=\textwidth]{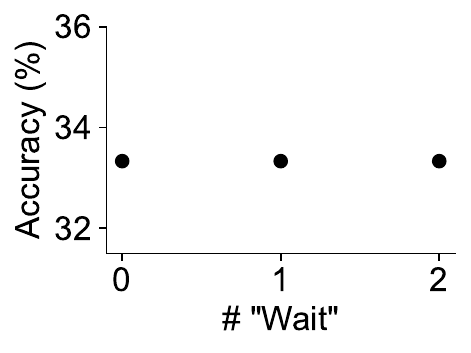}
        \caption{Deepseek-V3}
        
    \end{subfigure}
    \hfill
    \begin{subfigure}[b]{0.24\textwidth}
        \includegraphics[width=\textwidth]{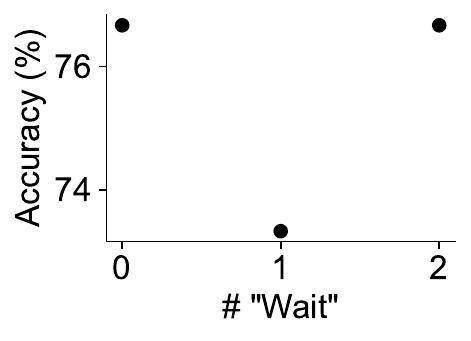}
        \caption{r1-distill-Qwen-32B}
        
    \end{subfigure}
    \hfill
    \begin{subfigure}[b]{0.24\textwidth}
        \includegraphics[width=\textwidth]{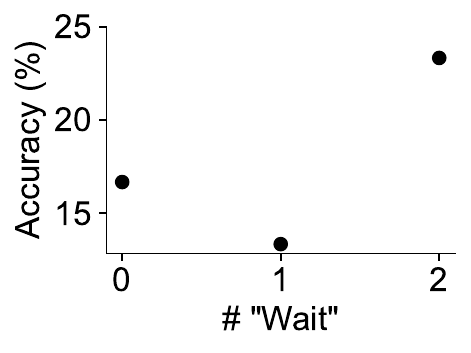}
        \caption{Qwen2.5-72B}
        
    \end{subfigure}
    \hfill
    \begin{subfigure}[b]{0.24\textwidth}
        \includegraphics[width=\textwidth]{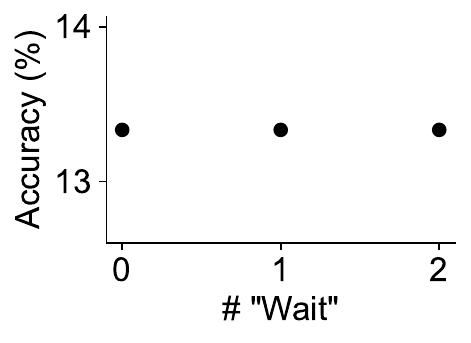}
        \caption{Qwen2.5-32B}
        
    \end{subfigure}
    
    \caption{Scaling up by appending "Wait" across five models reveals inconsistent performance in three models, which is caused by the model oscillating between answers on several problems. In contrast, the other models maintain steady performance.
}
    \label{fig:wait}
\end{figure}

\noindent\textbf{Results.} Figure \ref{fig:wait} illustrates that the performance of s1 fluctuates after the number of "Wait" instances exceeds four, which is due to the model changing its answer to a single problem repeatedly. Before reaching four "Wait" instances, the accuracy remains relatively steady, though the model alters its responses for two specific problems. For instance, when one "Wait" is appended, it only correctly solves the first problem, while with two "Wait" instances, it only solves the second problem. Therefore, simply appending "Wait" leads to inconsistent progress for s1. Additionally, we observe that for most problems, the answers remain unchanged.

Further experiments with the other four models reveal that performance remains steady for two models, while the other two models exhibit an initial decline followed by an increase. For the two models that maintain steady performance, the answers to all correctly solved problems remain consistent, while answers to a few incorrectly solved problems change, though none of these become correct. For the two models exhibiting fluctuations, r1-distill-Qwen-32B alternates between correct and incorrect answers for a single problem, while Qwen2.5-72B first turns one correct problem into an incorrect one after one "Wait" instance, then not only restores that problem to correct but also adjusts two other initially incorrect problems to correct after the second "Wait." For all models, both the answers and solutions exhibit minimal changes for the majority of the problems.

Overall, scaling up by appending "Wait" results in inconsistent performance—answers can either improve, worsen, or remain unchanged. As more "Wait" instances are appended, answers may alternate between correct and incorrect as shown in Figure \ref{fig:flip}. Therefore, without an oracle to determine when to stop appending, there is no guarantee of improvement.

\begin{figure}[t]
\begin{center}
\includegraphics[width=1\linewidth]{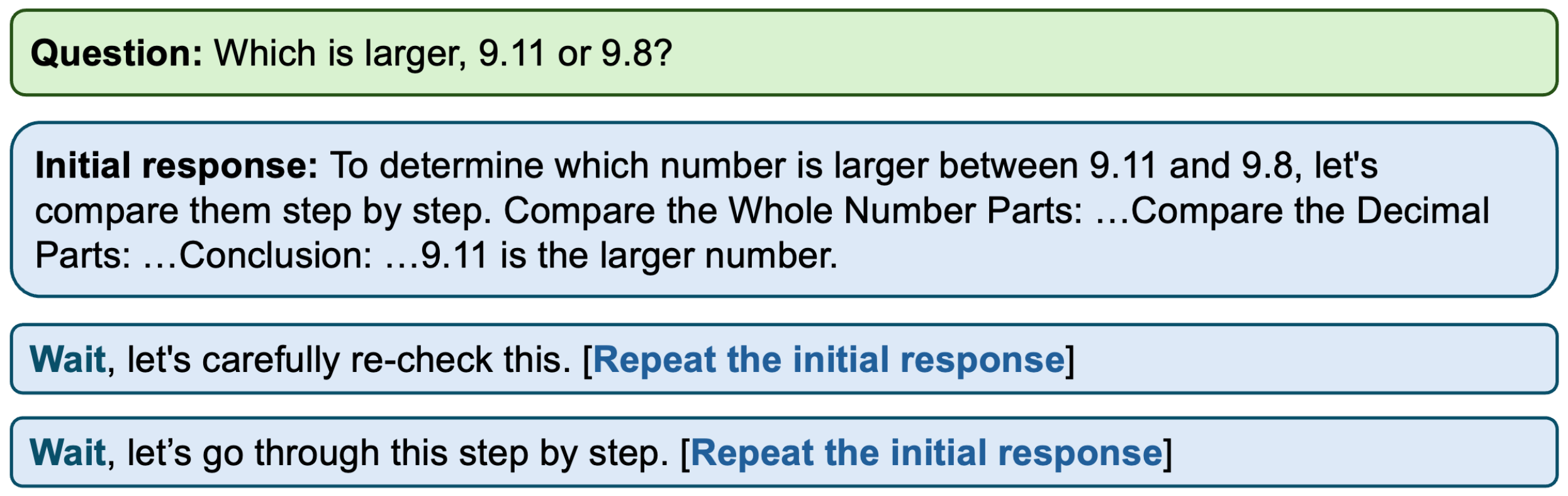}
\end{center}
\caption{An example of the model repeating the response after "Wait" is appended. The model often begins with a reflective phrase such as "Let's re-check this."} \label{fig:repeat}
\end{figure}

While some problems may benefit from additional computation, for the majority, the answers and solutions remain largely unchanged. As shown in Figure \ref{fig:repeat}, after appending "Wait," the models often simply repeat the initial response. Thus, we conclude that scaling up by appending "Wait" is inefficient. It occasionally leads to slight improvements, but for most problems, it results in repeated answers and similar solutions.

\subsubsection{Inefficiency Commonly Observed} \label{sec:repeat}
In the preceding section, we identified inefficiencies in the model's output, noting that appending "Wait" rarely altered the responses for the majority of problems. Through manual analysis, we observed that the model frequently either repeated the same response verbatim or generated a slightly modified version with minimal substantive changes. In this section, we further investigate this inefficiency.

\noindent\textbf{Setup.}  We focus on three key aspects in our analysis. First, we investigate how accuracy changes as additional instances of "Wait" are appended. Second, we evaluate how frequently the model's answer remains unchanged compared to previous responses. Specifically, we compare the answer generated after appending one "Wait" ("Wait"×1) to the initial answer, and then compare the answer from appending one "Wait" to the answer from appending two "Wait" ("Wait"×2). Finally, we assess how often the model repeats the same response verbatim after the second "Wait" ("Wait"×2) as it did after the first "Wait." Notably, we consider only exact matches, which provide a conservative measure of repetitive behavior.

For all experiments, we set the temperature to 0.7 across the four models and append two "Wait" instances after the initial response. Each experiment is conducted over three runs. Due to Deepseek-V3 producing nonsensical output in one of the three runs, we exclude that run from its evaluation. For all evaluated aspects, we calculate the average performance across all valid runs for each model.

\noindent\textbf{Results.} Table \ref{tab:repeat} reveals that accuracy improves slightly for three models, while it remains unchanged for one. Additionally, the answers for a significant portion of the problems show no change. Furthermore, we observe that some models frequently engage in what appears to be fake reflection—beginning responses with phrases such as "Let's re-check this"—only to repeat the same solution afterward. This behavior is most prevalent in Deepseek-V3, while 1-distill-Qwen-32B also exhibits a high repetition rate. 

These results highlight the inefficiency of scaling up responses by simply appending "Wait." For the majority of problems, the answers remain largely unchanged, and the responses are often just repetitions of the previous response, occasionally accompanied by superficial attempts at reflection. This underscores the limitations of this approach in generating meaningful or novel outputs.

\begin{table}[t]
\begin{center}
\renewcommand{\arraystretch}{1.2} 
\begin{tabular}{l|ccc|cc|c}
\toprule
\multirow{2}{*}{\bf Model} & \multicolumn{3}{c}{\bf Accuracy} & \multicolumn{2}{c}{\bf Answer} & \multicolumn{1}{c}{\bf Response} \\
\cmidrule(l){2-7} 
& Init& \small ``Wait''$\times1$& \small ``Wait''$\times2$& \small ``Wait''$\times1$& \small ``Wait''$\times2$& \small ``Wait''$\times2$\\ 
\midrule
\small Deepseek-V3 & 28.3 & 30 & 30 & 85 & 98.3 & 86.7 \\
\small Qwen2.5-32B-Instruct & 13.3 & 14.4 & 15.6 & 85.6 & 87.8 & 12\\
\small Qwen2.5-72B-Instruct& 15.6 & 16.7 & 18.9 & 70 & 82 & 1.1\\
\small r1-distill-Qwen-32B& 77.8& 77.8& 77.8 & 98.9 & 97.8 & 55.6\\
\bottomrule
\end{tabular}
\end{center}
\caption{Average accuracy and repetition rate of answers and responses from four models with a temperature of 0.7. "Init" represents the initial response, while "Wait"$\times$1 and "Wait"$\times$2 indicate the responses after the first and second "Wait" are appended. For the repetition rate, the responses are compared with their previous responses, meaning "Wait"$\times$1 corresponds to the results compared with the initial response.}\label{tab:repeat}
\end{table}

\subsubsection{Unsuccessful Attempt to Fix The Inefficency} \label{sec:fix}
As observed in the previous section, the responses generated by the models can be highly repetitive. To address this, we explore whether increasing the temperature can enhance the variability of the responses.

\noindent\textbf{Setup.} For Deepseek-V3, we set the temperature to 2, while for the other three models, we set it to 1.99, as their supported temperature range is limited to values below 2. We append two "Wait" instances after the initial response and conduct two runs. Once again, Deepseek-V3 produces nonsensical output in one of the two runs, so we exclude that run from its evaluation. 

We measure accuracy, the repetition rate of answer and response, as defined in Section \ref{sec:repeat}. The reported numbers represent the average across all valid runs for each model.

\noindent\textbf{Results.} Table \ref{tab:fix} demonstrates that while a high temperature significantly reduces the repetition rate of the responses, it also leads to a substantial degradation in accuracy for most models, with the exception of r1-distill-Qwen-32B, where the impact of high temperature is minimal. Additionally, for three out of the four models, a large proportion of the answers remain unchanged, with Qwen2.5-72B-Instruct being the only exception.

Overall, we find that while a high temperature can effectively reduce repetition rates in response, it often comes at the cost of reduced accuracy. Moreover, it does not significantly lower the repetition rate of answers in most cases. We also experimented with using a frequency penalty to mitigate repetition, but since the models frequently repeat entire responses rather than individual sentences, this approach proved largely ineffective.
\begin{table}[t]
\begin{center}
\renewcommand{\arraystretch}{1.2} 
\begin{tabular}{l|ccc|cc|c}
\toprule
\multirow{2}{*}{\bf Model} & \multicolumn{3}{c}{\bf Accuracy} & \multicolumn{2}{c}{\bf Answer} & \multicolumn{1}{c}{\bf Response} \\
\cmidrule(l){2-7} 
& Init& \small ``Wait''$\times1$& \small ``Wait''$\times2$& \small ``Wait''$\times1$& \small ``Wait''$\times2$& \small ``Wait''$\times2$\\ 
\midrule
Deepseek-V3& 20 & 23.3 & 20 & 70 & 80 & 3.33 \\

Qwen2.5-32B-Instruct& 0 & 1.7& 1.7& 72.8& 67.3 & 0\\

Qwen2.5-72B-Instruct& 1.7 & 3.3 & 8.3& 15& 26.7 & 0\\

r1-distill-Qwen-32B& 73.4& 76.7& 75& 96.7& 96.7 & 46.7\\
\bottomrule
\end{tabular}
\end{center}
\caption{Average accuracy and repetition rate of answers and responses from four models with a high temperature setting.}\label{tab:fix}
\end{table}
\begin{tcolorbox}[
    colback=blue!5!white, 
    colframe=black, 
    sharp corners=south, 
    title={\textbf{Takeaways for scaling up by appending ``Wait''}}, 
    fonttitle=\bfseries, 
    coltitle=white, 
    colbacktitle=black
]
When scaling up by appending "Wait," we observe inconsistent progress: performance may improve, decline, or remain steady. This inconsistency arises because models often oscillate in their responses to certain problems. Furthermore, we note that for the majority of problems, scaling up barely alters the answers, and the models frequently repeat previous responses. Despite exploring potential fixes, such as using a high temperature, we have yet to find a straightforward solution to address this inefficiency. This highlights the challenges in improving model performance through simple scaling mechanisms like appending "Wait."
\end{tcolorbox}

\section{Conclusion and Discussion}
In this work, we examine the key components of simple test-time scaling and find that the test-time scaling behavior can largely be attributed to scaling down by enforcing a maximum length. This constraint forces the model to provide answers prematurely, effectively imposing a manual upper limit on performance. Notably, whether the model has been fine-tuned on long CoT data or not makes no difference to this behavior. Additionally, scaling up by appending "Wait" leads to inconsistent progress and significant inefficiencies.

We argue that while simple test-time scaling exhibits patterns similar to those of o1-like models, there is a fundamental difference between the two. For o1-like models, the evaluation process allows them to develop mature solutions, typically utilizing the highest supported maximum length. These models naturally learn to scale up test-time compute through reinforcement learning (RL), achieving test-time scaling behavior. Behind the similar appearance of scaling behavior lies a crucial distinction: for o1-like models, the model is at its peak performance when the highest supported maximum length is allowed, and both performance and compute naturally increase as RL advances. In contrast, simple test-time scaling relies on a single model constrained by a manually imposed performance cap, leading to performance degradation as the maximum length is reduced.

Furthermore, we observe that long-CoT models exhibit significant redundancy. For many problems, only one-third of the tokens are necessary to reach the correct answer, while in some cases, the model may benefit from additional thinking time. These findings align with recent research \citep{chen2025think23overthinkingo1like,wang2025thoughtsplaceunderthinkingo1like} on efficient reasoning, highlighting the importance of developing adaptive scaling methods tailored to individual problems.




\bibliography{colm2025_conference}
\bibliographystyle{colm2025_conference}

\end{document}